\documentclass[10pt,twocolumn,letterpaper]{article}

\usepackage[pagenumbers]{cvpr}

\usepackage{graphicx}
\usepackage{amsmath}
\usepackage{amssymb}
\usepackage{booktabs}
\usepackage{url}
\usepackage{amsmath}
\usepackage{amssymb}
\usepackage{bm}
\usepackage{multirow}
\usepackage{pifont}
\usepackage[pagebackref,breaklinks,colorlinks]{hyperref}
\usepackage[accsupp]{axessibility}
\usepackage[capitalize]{cleveref}

\newcommand{\cmark}{\ding{51}}
\newcommand{\xmark}{\ding{55}}

\crefname{section}{Sec.}{Secs.}
\Crefname{section}{Section}{Sections}
\Crefname{table}{Table}{Tables}
\crefname{table}{Tab.}{Tabs.}

\begin{document}

\title{Prompt-Guided Zero-Shot Anomaly Action Recognition \\ using Pretrained Deep Skeleton Features}

\author{
Fumiaki Sato\thanks{~Equal contribution.}, ~~Ryo Hachiuma\footnotemark[1], ~~Taiki Sekii \\
Konica Minolta, Inc. \\
{\tt\small \{fumiaki.sato.jp,rhachiuma,taiki.sekii\}@gmail.com} \\
}

\maketitle

\begin{abstract}
This study investigates unsupervised anomaly action recognition, which identifies video-level abnormal-human-behavior events in an unsupervised manner without abnormal samples, and simultaneously addresses three limitations in the conventional skeleton-based approaches: target domain-dependent DNN training, robustness against skeleton errors, and a lack of normal samples.
We present a unified, user prompt-guided zero-shot learning framework using a target domain-independent skeleton feature extractor, which is pretrained on a large-scale action recognition dataset.
Particularly, during the training phase using normal samples, the method models the distribution of skeleton features of the normal actions while freezing the weights of the DNNs and estimates the anomaly score using this distribution in the inference phase.
Additionally, to increase robustness against skeleton errors, we introduce a DNN architecture inspired by a point cloud deep learning paradigm, which sparsely propagates the features between joints.
Furthermore, to prevent the unobserved normal actions from being misidentified as abnormal actions, we incorporate a similarity score between the user prompt embeddings and skeleton features aligned in the common space into the anomaly score, which indirectly supplements normal actions.
On two publicly available datasets, we conduct experiments to test the effectiveness of the proposed method with respect to abovementioned limitations.
\end{abstract}

\begin{figure}[t]
\centering
\includegraphics[clip, width=\hsize]{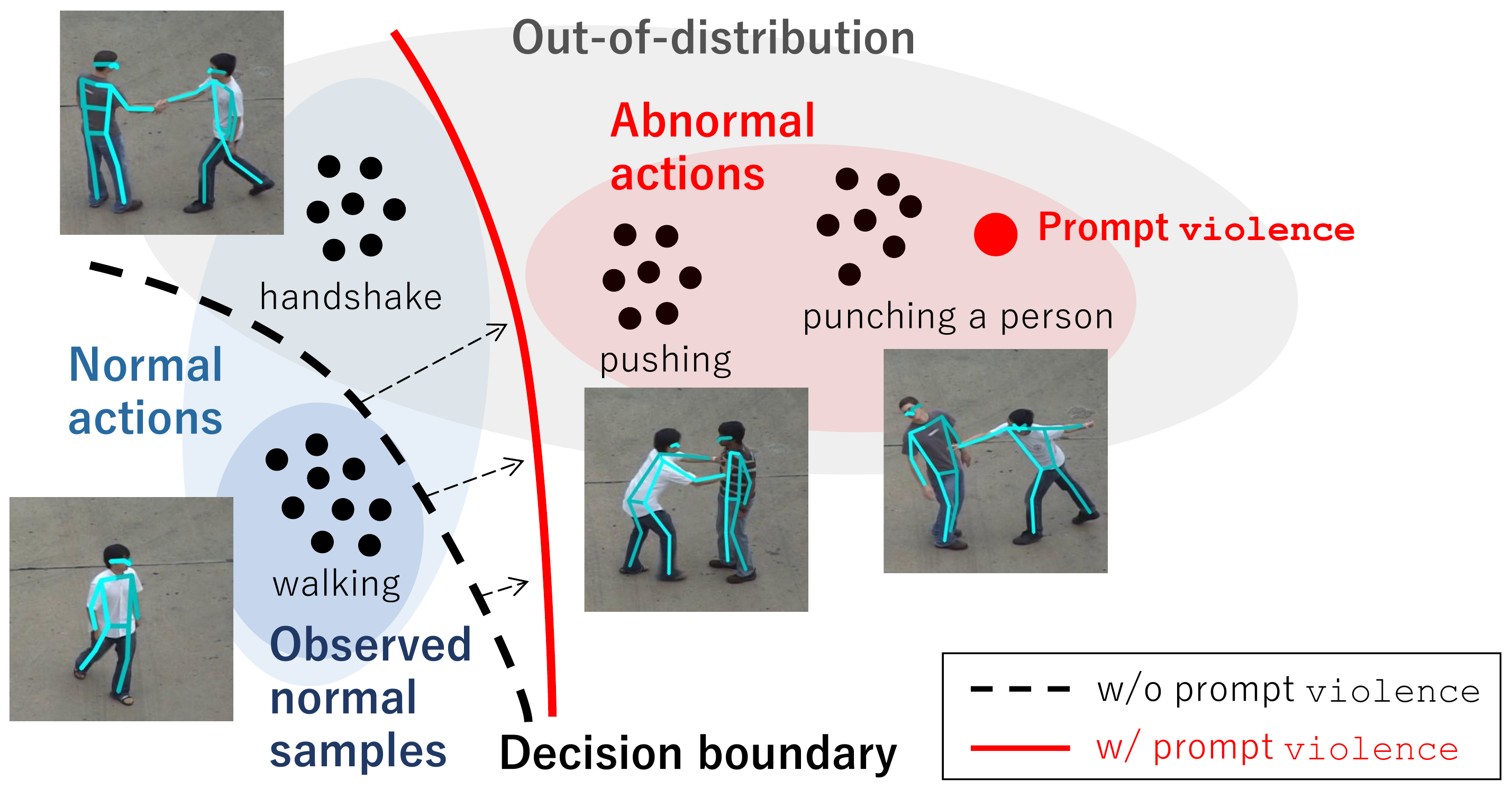}
\caption{Modeling the distribution of skeleton features per video for identifying violent action samples as abnormal while only walking samples are observed as normal samples in the training phase.
The decision boundary (black dotted line), which is learned from only normal samples, is shifted by the proposed prompt-guided anomaly score in the direction of embedding the prompt \texttt{violence} input by the user (red line).
Handshake samples unobserved during the training are out-of-distribution but normal and are incorrectly recognized as abnormal without such prompt due to a lack of normal samples.
However, by adding this prompt, handshake samples can be correctly identified as normal while walking samples are recognized as normal.}
\label{fig:t-sne}
\vspace{-3mm}
\end{figure}

\section{Introduction}
\label{sec:intro}
Anomaly action recognition, the task that detects whether the person(s) in the video is/are behaving abnormally~\cite{Su2020ECCV, Cheng2021ICPR, Islam2021IJCNN, Markovitz2020CVPR, Morais2019CVPR, Zaheer2022CVPR, Wang2022NNLS, Liu2022AS}, becomes an essential piece of technology for averting accidents and crimes~\cite{Cheng2021ICPR,Sultani2018CVPR}.
The previous work can be classified into two methods that leverage appearance information from the videos~\cite{Cheng2021ICPR, Islam2021IJCNN, Zaheer2022CVPR, Wang2022NNLS} or only their human skeletons~\cite{Su2020ECCV, Morais2019CVPR, Markovitz2020CVPR, Liu2022AS}.
With the help of Deep Neural Networks (DNNs), the earlier methods identify abnormal actions by analyzing appearance features from videos.
On the other hand, the latter methods use only low-information skeleton sequences extracted by applying multi-person pose estimation approaches~\cite{Cao2021PAMI, Fang2017ICCV, Sekii2018ECCV} (simply referred to as {\it pose detectors}) to the videos and thus are relatively robust to changes in the appearances of the person and background~\cite{Weinzaepfel2021IJCV}.

Furthermore, the previous methods identify the abnormal actions for each frame~\cite{Zaheer2022CVPR, Wang2022NNLS, Morais2019CVPR, Liu2022AS} or each video clip~\cite{Su2020ECCV, Cheng2021ICPR, Islam2021IJCNN, Markovitz2020CVPR}.
They also follow supervised~\cite{Su2020ECCV, Cheng2021ICPR, Islam2021IJCNN} or unsupervised~\cite{Zaheer2022CVPR, Wang2022NNLS, Morais2019CVPR, Markovitz2020CVPR, Liu2022AS} manners depending on whether annotations are given.
Because the annotation cost is lower, the proposed method makes use of the skeleton-based approach, which recognizes abnormal actions at the video level in an unsupervised manner.

This study uses two assumptions; the users can define the categories of abnormal actions\footnote{It is obvious that the normal actions can be defined as the opposite of the abnormal actions while we focus on the definition of the abnormal actions to simplify the explanation.} (\eg, \texttt{violence} in \cref{fig:t-sne}), and the observed training samples consist of normal actions.
Additionally, unobserved training actions are referred to as {\it out-of-distribution} (OoD) (\eg, ``handshake'' and ``pushing'' in \cref{fig:t-sne}).
The OoD actions include unobserved normal actions (\eg, only ``handshake'' in \cref{fig:t-sne}) when a sufficient variety of normal samples is not observed in the training phase.

This study focuses on limitations in the previous studies~\cite{Markovitz2020CVPR, Morais2019CVPR} to improve the scalability, such as expanding to different applications and enhancing the performance, as described below.

\noindent \textbf{Target domain-dependent DNN training.}
The previous methods require time for training the DNNs with expensive computational resources for each scene when the applications are initialized, or the domain shift between the training and inference phases occurs, such as changes in distribution over time.
As a result, there are limitations on the applications, and use restrictions are put in place.

\noindent \textbf{Lack of normal samples.}
In the real-world scenario, a variety of normal actions cannot be obtained to train the DNNs.
In such cases, most actions are regarded as abnormal, that is, normal but OoD actions are misidentified abnormal.
Consequently, it is preferred that users be able to define the target abnormal and/or normal actions to be recognized, as shown in \cref{fig:t-sne}.

\noindent \textbf{Robustness against skeleton errors.}
The majority of traditional skeleton-based methods~\cite{Sijie2018AAAI, Morais2019CVPR, Markovitz2020CVPR, Liu2022AS} presuppose that DNNs, like Graph Neural Networks (GNNs), propagates the features densely between joints. 
Thus, the anomaly recognition accuracy degrades if joint detection errors (False Positives (FPs) and False Negatives (FNs)) occur in the pose detection or if the multi-person pose tracking fails as a result of environmental noises, such as illumination fluctuations.

To simultaneously overcome these limitations, this paper proposes a novel, prompt-guided zero-shot\footnote{We consider that normal samples do not directly contribute to representation learning, and thus follow a definition of zero-shot learning by Xian \etal~\cite{Xian2018CVPR}, which directly obtains representations of unknown classes.} framework for recognizing abnormal actions using a pretrained deep feature extractor with human skeleton sequences input.
The method does not require observation of abnormal actions or their ground-truth labels to train DNNs.
In particular, to address the first training limitation, we model the distribution of normal samples during the training\footnote{{\it Training} refers to learning normal and/or abnormal samples and is distinguished from {\it pre-training} on a dataset for action recognition, which does not learn such samples.} phase by utilizing DNNs with skeleton feature representations that have been pretrained on a sizable action recognition dataset, such as Kinetics-400~\cite{Carreira2017CVPR}.
The weights of the skeleton feature extractor are frozen during the training phase, and thus their features are relatively independent to the targeted domain.

To address the second normal-sample limitation, we reduce the misdetections that the normal action is identified as abnormal by utilizing the text prompts regarding the abnormal actions provided by the users to indirectly supplement the information of the normal actions.
We integrate a similarity score between the skeleton features and the text embeddings extracted from a text encoder into the anomaly score.
By implementing a contrastive learning scheme between skeleton features and text embeddings, it can be accomplished in the context of vision and language, which has been actively studied in recent years.

Inspired by a point cloud deep learning paradigm, we introduce a more straightforward DNN that sparsely propagates the features between joints as such a feature extractor, improving the robustness against such skeleton errors in the third limitation mentioned above.
This architecture eliminates the constraints on input skeletons such as the input joint size and order, which are dependent on the dataset/domain.
It allows us to divert the pretrained feature extractor frozen across different domains/datasets without any fine-/hyperparameter-tuning and to simultaneously model both the distribution of normal samples and the joint-skeleton text embedding space over the domains/datasets.

In summary, the main contributions of this work are listed as follows:
(1) We demonstrate experimentally that DNN training with normal samples can be eliminated via the skeleton feature representations pretrained using a large-scale action recognition dataset.
(2) We show that the zero-shot learning paradigm, which handles the skeleton features and text embeddings in the common space, can be efficient for modeling the distributions of the normal and abnormal actions.
It is supported by a brand new unified framework that incorporates user guided text embeddings into the computation of the anomaly score.
(3) We demonstrate experimentally that the permutation-invariant architecture, which sparsely propagates the features between joints, works as the skeleton feature extractor that models the normal samples and the joint-skeleton text embedding space over domains and enhances robustness against skeleton errors.

\section{Related work}\label{sec:relatedwork}
\subsection{Video anomaly detection}\label{sec:videoanomalydet}
The video anomaly detection task identifies abnormal actions in relatively short time (frame-by-frame) intervals compared to the anomaly action recognition task, introduced in \cref{sec:anoamlyactrec}.
Early appearance-based methods use hand-crafted motion features as input, such as histograms of the pixel change~\cite{Benezeth2009CVPR} or the optical flow~\cite{Adam2008PAMI}.
Due to the DNNs' recent advancements, 3D Convolutional Neural Networks (CNNs) are now being used to extract spatio-temporal features in a data-driven fashion~\cite{Wang2019ICCV, Cheng2021ICPR, Zaheer2022CVPR}.
On the other hand, the skeleton-based approaches~\cite{Morais2019CVPR, Liu2022AS, Matei2022Arxiv} concentrate on the DNN architecture, such as recurrent neural networks~\cite{Morais2019CVPR, Matei2022Arxiv} or GNNs~\cite{Liu2022AS}, to model the motion features from input human skeleton sequences.
Our approach makes use of the advantages of skeleton-based approaches, which are more resistant to changes in a person's appearance or background as a result of training~\cite{Weinzaepfel2021IJCV}.

The skeleton-based video anomaly detection can be classified into supervised~\cite{Matei2022Arxiv} and unsupervised learning approaches~\cite{Morais2019CVPR, Liu2022AS}.
The latter approaches~\cite{Morais2019CVPR, Liu2022AS} identify the abnormal actions under the assumption that normal actions can be observed regularly, and such data can be gathered easily.
These methods do not require manual labeling of the training dataset.
Comparing the observed and reconstructed human skeleton sequences during the inference phase allows them to identify abnormal behavior.

\subsection{Anomaly action recognition}\label{sec:anoamlyactrec}
The anomaly action recognition task can identify video-level abnormal actions that consist of intermittent actions in relatively long time intervals, compared to the video anomaly detection task.
Due to the advantage of having only a few restrictions against the target's abnormal actions, this paper takes on this task.
Anomaly action recognition can also be classified into supervised and unsupervised learning contexts, as same as \cref{sec:videoanomalydet}.
In the supervised context, the appearance-based approaches apply the 3D CNNs to RGB and optical flow images~\cite{Cheng2021ICPR}, or do the Long-Short Term Memory networks to the outcomes of the background/frame subtraction algorithms~\cite{Islam2021IJCNN}.
On the other hand, in the unsupervised context, the skeleton-based approach~\cite{Markovitz2020CVPR} uses the reconstructed human skeleton sequences from the observation, similar to the video anomaly detection task.
The unsupervised skeleton-based methods have the limitations listed in \cref{sec:intro} for tasks like video anomaly detection and anomaly action recognition.

\subsection{Zero-shot action recognition}
The field of vision and language has been actively researching the zero-shot visual recognition task, which identifies the unseen target in the visual data with a text prompt that describes the target, as a result of the field's rapid advancement in natural language processing.
For instance, the zero-shot image classification task~\cite{Radford2021ICML,Changpinyo2021CVPR} takes a pair of images and its text prompt to recognize the category unseen during the training.
Also, the Visual Question Answering task~\cite{Cascante-Bonilla2022CVPR, Gupta2022CVPR} takes an input of a pair of images and its corresponding question via the text.
The performance of such tasks is significantly enhanced by introducing the contrastive learning~\cite{Radford2021ICML} between image features and text embeddings extracted from the prompts.

Recently, contrastive learning has also been introduced~\cite{Tevet2022ECCV, Nag2022ECCV} to action recognition which takes advantage of the text prompts of an unseen target action.
In these approaches, the action is recognized in a zero-shot manner that aligns the text embedding and the appearance or skeleton feature extracted from the video during the training.
This study introduces the zero-shot method in the context of the task of identifying abnormal actions to enhance the modeling of the distribution of abnormal actions.

\subsection{Skeleton-based action recognition}
A supervised anomaly action recognition task can be treated as a supervised action recognition task that uses the dataset with normal and abnormal ground-truth labels.
The relationships between time-series joints have been studied using a variety of skeleton-based methods~\cite{Sijie2018AAAI, Shi2019CVPR, Liu2020CVPR, Chi2022CVPR} that primarily use GNNs.
In contrast, SPIL~\cite{Su2020ECCV} treats human skeleton sequences as an input 3D point cloud and is a technique that competes with the proposed method only to the architectural concept.
It models the dense relationships between the joints by an attention mechanism~\cite{Vaswani2017NeurIPS}.
The proposed architecture improves robustness against input errors, such as FP and FN joints or pose tracking errors, by sparsely propagating the features between the joints.

\section{Method}\label{sec:method}
\begin{figure}[t]
\centering
\includegraphics[width=.98\linewidth]{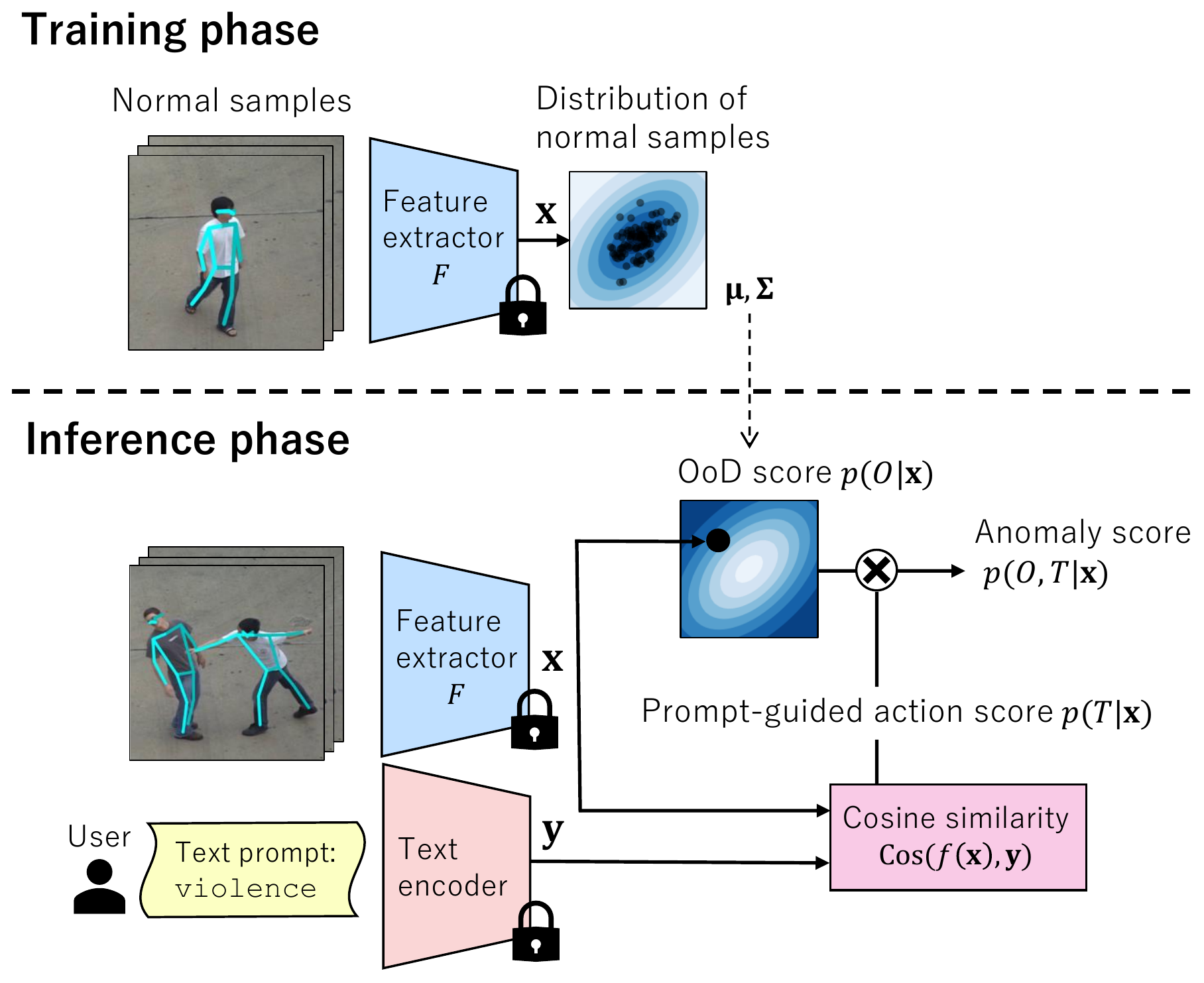}
\caption{Overview of the proposed framework. DNN pretraining not included.}
\label{fig:overview}
\vspace{-3mm}
\end{figure}
\begin{figure}[t]
\centering
\includegraphics[width=.98\linewidth]{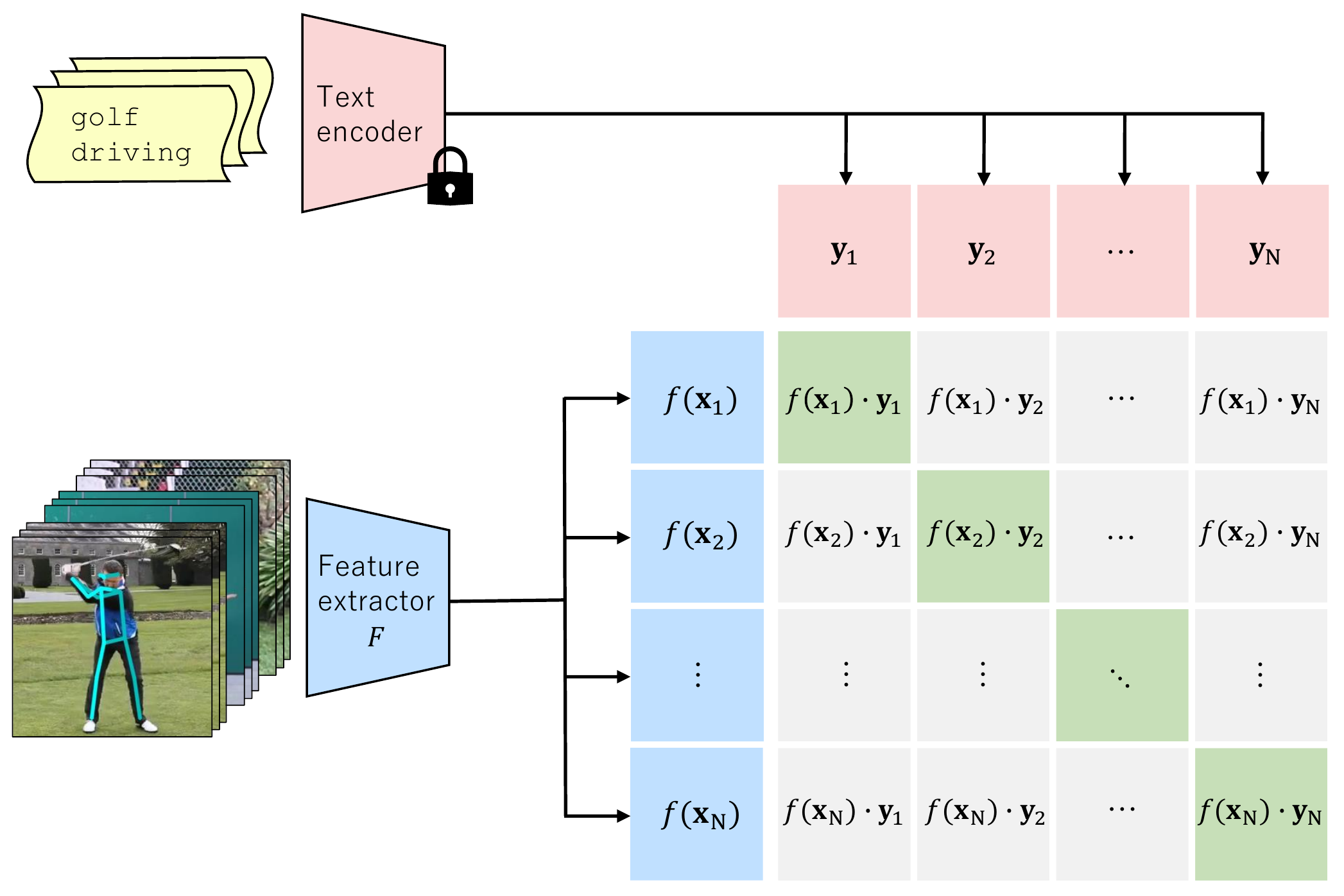}
\caption{Overview of the contrastive learning between the skeleton features and the text embeddings in the pretraining phase.}
\label{fig:pretrain}
\vspace{-3mm}
\end{figure}
The pipeline of the framework consists of (1) pretraining, wherein the DNNs are trained on an action recognition dataset without normal samples; (2) training, wherein only the distribution of normal samples is computed (trained), while no DNNs are trained; and (3) inference, wherein the anomaly score is computed using distribution and the text prompts of an unseen action.
\cref{fig:overview} illustrates steps (2) and (3) in the target domain.
The pretraining phase is described in \cref{sec:training}.

First, in both the training and inference phases, the multi-person pose estimation is applied to the input video for extracting the human joints.
Then, each joint is transformed into an input vector $\mathbf{v}$ for the DNNs.
$\mathbf{v}$ is a seven-dimensional vector consisting of the two-dimensional joint coordinates on the image, the time index, the joint confidence, the joint index, and the two-dimensional centroid coordinates calculated from the human joints.
Each element in the input vector is normalized between $0$ and $1$.
All input vectors $\mathcal{V}=\{\mathbf{v}_1, \cdots ,\mathbf{v}_J\}$ are treated as a 3D point cloud, input to the DNNs to extract the skeleton feature $\mathbf{x}  \in \mathbb{R}^{S}$.
The anomaly score is defined as the joint probability of the probability $p(O|\mathbf{x})$ that represents $\mathbf{x}$ does not belong to the normal samples and the probability $p(T|\mathbf{x})$ that represents $\mathbf{x}$ includes the abnormal actions specified by the user and is expressed as follows:
\begin{equation}\label{eq:joint}
p(O,T|\mathbf{x})=p(O|\mathbf{x})p(T|\mathbf{x}),
\end{equation}
where $O$ and $T$ are binary random variables.
In the following sections, each term on the right-hand side in \cref{eq:joint} and the training schema are described in detail.

In the training phase on the normal samples, the parameters for $p(O|\mathbf{x})$ model the distribution of $\mathbf{x}$ in the training samples.
The parameters for $p(T|\mathbf{x})$ are the text embeddings compared with $\mathbf{x}$ and are described in \cref{sec:user}.
We present a mechanism based on PointNet~\cite{Qi2017CVPR} to the feature extractor described in \cref{sec:FeatExt}, which is pretrained using a large-scale action recognition dataset, such as Kinetics-400.
As part of this pretraining phase, we introduce the contrastive learning scheme between skeleton features and text embeddings and train the DNNs using the action classification and contrastive losses, as described in \cref{sec:training}.
The following section goes into more detail about the aforementioned and the pretraining scheme.

\subsection{OoD score}\label{sec:anomal}
We approximate $p(O|\mathbf{x})$ in \cref{eq:joint} as a score called {\it OoD score}, which denotes $\mathbf{x}$ is not a normal sample, using the Mahalanobis distance, as follows:
\begin{equation}
    p(O|\mathbf{x})\sim \mathrm{min} \left(1.0, w_1\sqrt{(\mathbf{x}-\bm{\mu})^\intercal\bm{\Sigma}^{-1}(\mathbf{x}-\bm{\mu})}^{\frac{1}{w_2}} \right),
\end{equation}
where $(w_1,w_2)$ are a normalizing constant and a temperature parameter, respectively.
$\bm{\mu}$ and $\bm{\Sigma}$ are a mean vector and a covariance matrix of the distribution of training samples, respectively.

In the context of unsupervised image anomaly detection, Rippel~\etal~\cite{Rippel2021ICPR} modeled the anomaly score using the multivariate Gaussian distribution of the image features extracted from the normal samples while freezing the weights of the DNNs during the training phase.
In contrast to Rippel~\etal~\cite{Rippel2021ICPR} who concentrated on the image input, anomaly action recognition has to treat unordered input data of the human skeleton sequences, which include FPs and FNs of joints, pose tracking errors, or the change in the number of people, as described in \cref{sec:intro}.
The proposed feature extractor is built on PointNet~\cite{Qi2017CVPR}, which can handle a wide range of skeleton sequences because it has the permutation-invariant property for the order of input vectors.
In the experiments, we demonstrate that a case using only $p(O|\mathbf{x})$ as the anomaly score can also achieve unsupervised anomaly action recognition without updating the weights of the DNNs during the training phase.

\subsection{Prompt-guided action score}\label{sec:user}
We approximate $p(T|\mathbf{x})$ in \cref{eq:joint} as a score called {\it prompt-guided action score}, which denotes $\mathbf{x}$ includes the actions specified by the user.
In the inference phase, given $P$ text embeddings $\mathcal{Y}=\{\mathbf{y}_1, \cdots ,\mathbf{y}_P\}$ extracted by a text encoder, $p(T|\mathbf{x})$ is approximated as:
\begin{equation}
    p(T|\mathbf{x})\sim \mathrm{min} \left(1.0, w_1\mathrm{PromptScore}(\mathbf{x}|\mathcal{Y})^{\frac{1}{w_2}}\right).
\end{equation}
$\mathrm{PromptScore}(\cdot|\cdot)$ is formulated as:
\begin{equation}\label{eq:PAS}
\begin{aligned}
&\mathrm{PromptScore}(\mathbf{x}|\mathcal{Y})\\
&=\mathrm{max}\left(\mathrm{Cos}\left(f(\mathbf{x}), \mathbf{y}_1\right), \cdots , \mathrm{Cos}\left(f(\mathbf{x}), \mathbf{y}_P\right)\right),
\end{aligned}
\end{equation}
where $\mathrm{Cos}(\cdot,\cdot)$ represents the cosine similarity between two vectors, and $f$ denotes pretrained multilayer perceptron (MLP) to align the dimension of $\mathbf{x}$ and $\mathbf{y}$.

\subsection{Pretraining}\label{sec:training}
This section discusses the proposed pretraining scheme using a large-scale action recognition dataset.
We use contrastive learning between the skeleton features and the text embeddings extracted from action class names in the pretraining phase as well as multi-task learning on the action classification task, which uses the video-level action labels.
We define the total loss $\mathcal{L}$ that consists of the action classification loss $\mathcal{L}_\text{cls}$ and the contrastive loss $\mathcal{L}_\text{cont}$ in a batch of $N$ videos as follows:
\begin{equation}
  \mathcal{L} = \alpha \sum_{i=1}^{N} \mathcal{L}_{\text{cls},i} + (1 - \alpha) \mathcal{L}_\text{cont},
\end{equation}
where $\alpha$ is the mixing ratio of the loss functions.
The classification loss $\mathcal{L}_\text{cls}$ is formulated as the cross-entropy loss as follows:
\begin{equation}
    \mathcal{L}_\text{cls} = -\sum_{i=1}^{C}h_i\log\frac{\exp(l_i)}{\sum_{j=1}^{C}{\exp(l_j)}}, 
\end{equation}
where $C$ is the number of action classes, $(h_1,\cdots,h_C)$ is a ground-truth, one-hot action class vector, and $\left(l_1,\cdots,l_C\right)$ is a logit calculated from $\mathbf{x}$ using the Fully-Connected layers.

Based on the loss function proposed by CLIP~\cite{Radford2021ICML}, the contrastive loss $\mathcal{L}_\text{cont}$ is formulated using the symmetric contrastive loss, as follows:
\begin{equation}
  \mathcal{L}_\text{cont} = \frac{1}{2}\left(\mathcal{L}_\text{s2t} + \mathcal{L}_\text{t2s}\right),
\end{equation}
where $\mathcal{L}_\text{s2t}$ is the contrastive loss for the skeleton features against the text embeddings in the batch, and $\mathcal{L}_\text{t2s}$ is the loss which is the opposite with respect to $\mathcal{L}_\text{s2t}$~\cite{Ma2022CVPR}.
As illustrated in \cref{fig:pretrain}, the minimization of $\mathcal{L}_\text{s2t}$ and $\mathcal{L}_\text{t2s}$ maximizes the cosine similarity of the positive pairs of the skeleton feature and its action class text embedding.
Also, it minimizes the similarity of the negative pairs.
$\mathcal{L}_\text{s2t}$ and $\mathcal{L}_\text{t2s}$ are formulated as:
\begin{equation}
    \mathcal{L}_\text{s2t} = - \sum_{i=1}^{N} \log \frac{\exp(\mathrm{Cos}(f(\mathbf{x}_i), \mathbf{y}_i)/\tau)}{\sum_{j=1}^{N}\exp(\mathrm{Cos}(f(\mathbf{x}_i), \mathbf{y}_j)/\tau)},
\end{equation}
\begin{equation}
    \mathcal{L}_\text{t2s} = - \sum_{i=1}^{N} \log \frac{\exp(\mathrm{Cos}(f(\mathbf{x}_i), \mathbf{y}_i)/\tau)}{\sum_{j=1}^{N}\exp(\mathrm{Cos}(f(\mathbf{x}_j), \mathbf{y}_i)/\tau)},
\end{equation}
where a positive pair of $\mathbf{x}_i$ and its action class text embedding $\mathbf{y}_i$ is obtained from each video $i$.
$\tau$ is the learnable temperature parameter.

\subsection{Skeleton feature extractor}\label{sec:FeatExt}
\begin{figure}[t]
\centering
\includegraphics[width=.98\linewidth]{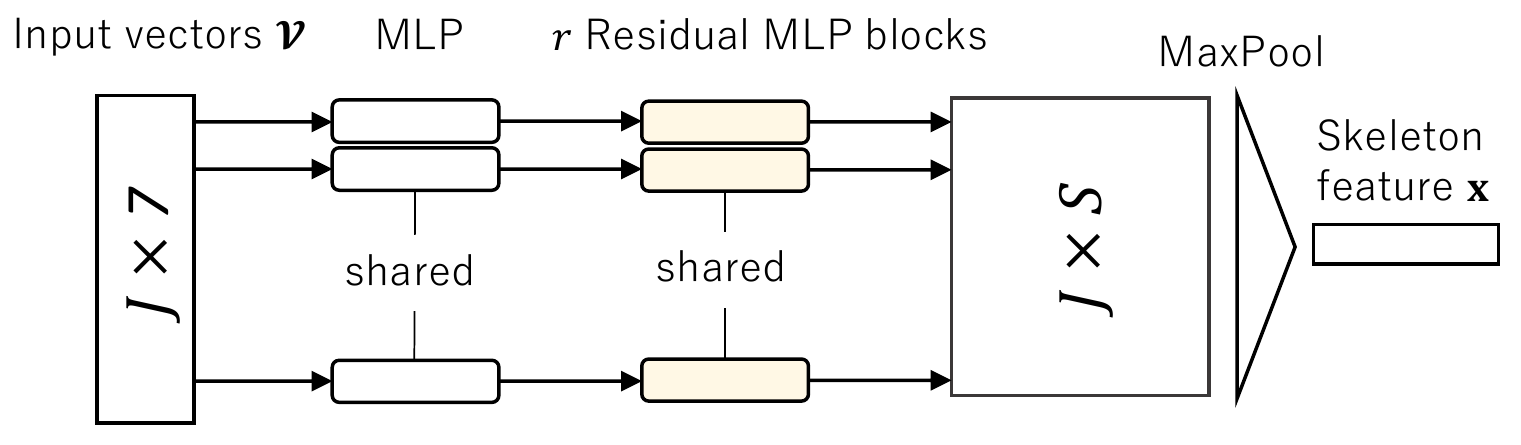}
\caption{The DNN architecture of the skeleton feature extractor.}
\label{fig:dnn_arch}
\vspace{-3mm}
\end{figure}
In this study, we design the skeleton feature extractor as a permutation-invariant DNN architecture that sparsely propagates the features between the joints leveraging the {\it Max-Pooling} operation to enhance the robustness described in \cref{sec:intro}, inspired by PointNet~\cite{Qi2017CVPR}.
This type of sparse feature propagation loosens the restrictions on the size or order of the input joints and can handle unordered skeleton sequences that include FPs and FNs of joints, pose tracking errors, or an arbitrary number of persons.

The architecture is shown in \cref{fig:dnn_arch}.
It is inspired by ResNet~\cite{He2016CVPR} and has a simple design composed of point-wise residual modules, which repeats the MLP for each joint.
Given the input vectors $\mathcal{V}=\{\mathbf{v}_1,\cdots ,\mathbf{v}_J\}$, we compute the skeleton feature $\mathbf{x}$ as follows:
\begin{equation}
    \mathbf{x} = F(\mathcal{V}) = \mathrm{MaxPool}\left(G\left(\mathbf{v}_1\right), \cdots ,G\left(\mathbf{v}_J\right)\right),
\end{equation}
where $\mathrm{MaxPool}(\cdot)$ is the symmetric operation of taking the maximum value for each channel from the input vectors.
$G$ is the DNNs that extract the high-order representation for each input joint.

In particular, $G$ first applies the MLP operation to the input vector before iteratively performing a residual MLP block $r$ times.
This residual MLP block extracts the output vector $\mathbf{u}_\text{out} \in \mathbb{R}^{D_\text{out}}$ from the input vector $\mathbf{u}_\text{in} \in \mathbb{R}^{D_\text{in}}$, which is formulated as:
\begin{equation}\label{eq:residualfunc}
    \mathbf{u}_\text{out}=\left\{\begin{matrix}
        \sigma\left(\phi\left(\mathbf{u}_\text{in}\right) + \mathbf{u}_\text{in}\right) & \text{if } D_\text{in} = D_\text{out} \\
        \sigma\left(\phi\left(\mathbf{u}_\text{in}\right) + \mathbf{W}_1\mathbf{u}_\text{in}\right) & \text{if } D_\text{in} \neq D_\text{out},
    \end{matrix}\right.
\end{equation}
where $\mathbf{W}_1 \in \mathbb{R}^{D_\text{out} \times D_\text{in}}$ is the learnable weight matrix.
Here, for presenting the bottleneck architecture into this residual block, we define $\phi$ as 3-layer MLPs as follows:
\begin{equation}
\begin{aligned}
    &\phi(\mathbf{u}_\text{in}) \\
    &= \mathrm{Norm}\left(\mathbf{W}_4 \cdot \sigma\left(\mathrm{Norm}\left(\mathbf{W}_3 \cdot \sigma\left(\mathrm{Norm}\left(\mathbf{W}_2\mathbf{u}_\text{in}\right)\right)\right)\right) \right),
\end{aligned}
\end{equation}
where $\mathbf{W}_2 \in \mathbb{R}^{\beta D_\text{out} \times D_\text{in} } $, $\mathbf{W}_3 \in \mathbb{R}^{ \beta D_\text{out} \times \beta D_\text{out} }$, and $\mathbf{W}_4 \in \mathbb{R}^{D_\text{out} \times \beta D_\text{out}}$ are the learnable weight matrices, and $\beta$ is the MLP bottleneck ratio.
$\mathrm{Norm}(\cdot)$ is the normalization layer, and $\sigma$ is the nonlinear activation function.

\section{Experiments}
By contrasting the accuracy with the traditional approaches in two settings, we assess the effectiveness of the proposed framework for the limitations described in \cref{sec:intro}.
One is that the abnormal action can be specified by the user.
The other is that its definition is ambiguous, leading the user to only describe a limited number of normal actions seen during the training phase.
These cases are respectively evaluated using two action recognition datasets, RWF-2000~\cite{Cheng2021ICPR} and Kinetics-250~\cite{Markovitz2020CVPR}.
Additionally, the ablation study verifies the proposed method's precise performance, including its robustness against skeleton detection errors, text prompt variation, and the domain shift.
The qualitative results using the UT-Interaction dataset~\cite{Ryoo2009ICCV} are shown in \cref{fig:overview}.
See the supplementary material for the implementation detail.

\subsection{Datasets}\label{sec:dataset}
Two action recognition datasets, RWF-2000~\cite{Cheng2021ICPR} and Kinetics-250~\cite{Markovitz2020CVPR}, are used for two evaluation settings discussed in \cref{sec:eval_set}.
Each dataset has been examined using the supervised learning-based (SL) and unsupervised learning-based (USL) approaches, respectively.
Note that our approach does not require any DNN training using normal samples, unlike such methods.
Furthermore, we use two large-scale action recognition datasets, Kinetics-400~\cite{Carreira2017CVPR} and NTU RGB+D 120~\cite{Liu2020PAMI}, for pretraining the proposed DNNs.
Each pretraining dataset is chosen respectively, taking into account the difference of the video source from the corresponding evaluation dataset or their domain gaps exist~\cite{Liu2020PAMI}, and a larger amount of actions is observed.
\cref{tab:Dataset} depicts the combination of datasets used during the evaluation (training and testing) and pretraining phases.

\noindent \textbf{Kinetics-400.} 
Kinetics-400~\cite{Carreira2017CVPR} is a large-scale action recognition dataset gathered from YouTube\footnote{\url{youtube.com}} videos with 400 action classes.
It contains 250K training and 19K validation 10-second video clips with 30 fps.

\noindent \textbf{RWF-2000.} RWF-2000~\cite{Cheng2021ICPR} is the violence recognition dataset gathered from YouTube videos.
The videos feature two actions, either violent or non-violent, captured by security cameras with a variety of people and backgrounds.
There are 1.6K training and 0.4K test 5-second video clips with 30 fps.
Two-class labels are annotated for each video.

\noindent \textbf{NTU RGB+D 120.} NTU RGB+D 120~\cite{Liu2020PAMI} is a large-scale action recognition dataset comprising videos captured in the laboratory environment.
It contains 114k videos with 120 action classes.
We use Cross-Setup (X-set) setting for the data split, where the camera setups are different between the training and testing phase~\cite{Markovitz2020CVPR}.

\noindent \textbf{Kinetics-250.} Kinetics-250~\cite{Markovitz2020CVPR} is a subset of the Kinetics-400 dataset and consists of videos with 250 action categories.
Since the Kinetics-400 dataset contains videos focused on human heads and arms, the accuracy of a skeleton-based approach is significantly impacted by these videos.
As a result, Markovitz~\etal~\cite{Markovitz2020CVPR} chose for evaluation the videos with the 250 action categories that performed the best in terms of action classification accuracy and allowed for the accurate detection of the skeleton.
In this study, we adopt the evaluation setting proposed by Markovitz~\etal, which is described in \cref{sec:eval_set}.

\subsection{Pose detectors}\label{sec:pose_detectors}
\noindent \textbf{PPN.}
As illustrated in~\cref{tab:Dataset}, in the experiments on the RWF-2000 dataset, we use the low-performance Pose Proposal Networks (PPN) detector~\cite{Sekii2018ECCV} under conditions of similar anomaly action recognition accuracy against several baselines (PointNet++ and DGCNN) because there are no publicly available skeleton data.
PPN~\cite{Sekii2018ECCV} detects human skeletons in a {\it Bottom-Up} manner from an RGB image at high speed.
They are made up of Pelee backbone~\cite{Wang2018NeurIPS} and trained on the MS-COCO dataset~\cite{Lin2014ECCV}.
The definition of the human skeleton is the same as OpenPose~~\cite{Cao2021PAMI}.
As input to the PPN, we resize the image to $320 \times 224$~$\text{px}^2$.

\noindent \textbf{HRNet.}
HRNet~\cite{Sun2019CVPR} is a {\it Top-Down} pose detector.
It obtains superior accuracy, while the computational cost includes a human detector (Faster R-CNN~\cite{Ren2015NeurIPS}) is expensive.
In the experiments on the Kinetics-250 dataset, we employ publicly available HRNet skeletons\footnote{\url{github.com/kennymckormick/pyskl}} given by Haodong~\etal~\cite{Duan2022CVPR}.

\subsection{Evaluation settings}\label{sec:eval_set}
\noindent \textbf{RWF-2000.}
In previous studies, the RWF-2000 dataset is used to assess violence action recognition accuracy of the models trained in a supervised manner.
In this paper, non-violence, and violent actions are defined as normal and abnormal, respectively.
The proposed method differs from supervised approaches in that the training phase of the proposed method uses non-violence action samples, and the DNN weights are frozen throughout this phase.
As a result, the proposed method recognizes the violence action in a zero-shot manner which does not require any observation of abnormal (violence) actions or the ground-truth labels during the training.
With five different hand-crafted text prompts that express the violence action, we test the proposed method's accuracy and use the one with the highest accuracy (see \cref{tab:CompText}).
The classification accuracy of violence or non-violence is used as the evaluation metric.
The pose detection average precision of the PPN is $36.4\%$ on the MS-COCO validation set.
Note that the baselines in the experiments use the highly accurate pose detector RMPE~\cite{Fang2017ICCV}, whose pose detection average precision is $72.3\%$.

\noindent \textbf{Kinetics-250.}
The evaluation setting on the Kinetics-250 dataset follows the previous study~\cite{Markovitz2020CVPR}.
In particular, we use the {\it Few vs. Many} setting that defines three to five action classes as normal and the rest of the action classes as abnormal.
In contrast to the other setting, where only a small number of classes are defined as abnormal, this one presents a greater challenge for the proposed method.
Two data splits, {\it random} and {\it meaningful} splits, are used for the evaluation.
{\it Few} classes at the {\it random} split consist of sets of three to five action classes which are randomly selected from the action classes defined in the Kinetics-250.
The {\it meaningful} split consists of sets of classes Markovitz~\etal subjectively grouped following some binding logic regarding the action's physical or environmental properties.
We employ the mean ROC-AUC for each split as the evaluation metric.

As previously mentioned, the proposed method only uses the label texts of {\it Few} classes as text prompts.
As a result, to determine the prompt-guided action score using such prompts, we update the definition explained in \cref{sec:CompSup} as the abnormal actions are conditioned.
The following is the modified \cref{eq:PAS}:
\begin{equation}
    \begin{aligned}
        &\mathrm{PromptScore}(\mathbf{x}|\mathcal{Y})\\
        &= 1 - \mathrm{max}\left(\mathrm{Cos}(f(\mathbf{x}), \mathbf{y}_1), \cdots , \mathrm{Cos}(f(\mathbf{x}), \mathbf{y}_P)\right).
    \end{aligned}
\end{equation}

\subsection{Comparisons with SoTA approaches} \label{sec:CompSup}

\begin{table}[tb]
\caption{Combination of datasets for evaluation of our method.}
\vspace{-3mm}
\centering
\scalebox{0.775}{
\begin{tabular}{c|c|c|c} \hline
\multirow{2}{*}{\shortstack{\\ Evaluation}} & \multirow{2}{*}{\shortstack{\\ Pretraining }} & \multirow{2}{*}{\shortstack{\\ Pose \\ Det}} & \multirow{2}{*}{\shortstack{\\ Baseline}} \\ 
    & & & \\ \hline
RWF-2000 & Kinetics-400 & PPN & SL \\
Kinetics-250 & NTU RGB+D 120 & HRNet & USL \\ \hline
\end{tabular}
}
\label{tab:Dataset}
\end{table}
\begin{table}[tb]
\caption{The performance comparison of the skeleton-based anomaly action recognition methods on the RWF-2000 dataset. The previous methods are trained in a supervised manner. *: HRNet skeletons are used as inputs. \dag: StructPool~\cite{Hachiuma2023CVPR} is employed as the network architecture.}
\vspace{-3mm}
\centering
\scalebox{0.775}{
\begin{tabular}{c|c|cccc} \hline
\multirow{2}{*}{Method} & \multirow{2}{*}{Acc. (\%)} & \multirow{2}{*}{Supervision} & DNN training \\ 
&  &  & in target dom. \\ \hline
PointNet++~\cite{Su2020ECCV} & $78.2$ & \multirow{4}{*}{\cmark} & \multirow{4}{*}{\cmark} \\ 
DGCNN~\cite{Su2020ECCV} & $80.6$ & & \\ 
SPIL~\cite{Su2020ECCV} & $89.3$ & & \\
ST-GCN~\cite{Sijie2018AAAI}* & $83.3$ & & \\ \hline
Only OoD & $71.8$ & \multirow{4}{*}{\xmark} & \multirow{4}{*}{\xmark} \\
Only prompt & $80.0$ & & \\
Ours & $82.5$ & & \\ 
Ours*\dag & $90.3$ & & \\ \hline
\end{tabular}
}
\label{tab:RWF2000Acc}
\end{table}
\begin{table}[tb]
\caption{The performance comparison of the skeleton-based anomaly action recognition methods on the Kinetics-250 dataset. The previous methods are trained in an unsupervised manner. \dag: StructPool~\cite{Hachiuma2023CVPR} is employed as the network architecture.}
\vspace{-3mm}
\centering
\scalebox{0.775}{
\begin{tabular}{c|cc|cc} \hline
\multirow{2}{*}{Method} & \multicolumn{2}{c|}{ROC-AUC} & \multirow{2}{*}{\shortstack{\\ Super- \\ vision}} & DNN training \\ 
& Random & Meaningful & & in target dom. \\ \hline
Learning Reg.~\cite{Markovitz2020CVPR} & $0.57$ & $0.59$ & \multirow{2}{*}{\xmark} & \multirow{2}{*}{\cmark} \\ 
Pose Clust.~\cite{Markovitz2020CVPR} & $0.65$ & $0.73$ & & \\ \hline
Only OoD & $0.68$ & $0.77$ & \multirow{4}{*}{\xmark} & \multirow{4}{*}{\xmark} \\
Only prompt & $0.52$ & $0.60$ & & \\
Ours & $0.69$ & $0.79$  & & \\
Ours\dag & $0.69$ & $0.78$  & & \\ \hline
\end{tabular}
}
\label{tab:K250Acc}
\end{table}

\begin{table*}[tb]
\begin{minipage}[r]{0.34\textwidth}
\caption{Comparison of the robustness against skeleton detection errors on the RWF-2000 dataset.}
\vspace{-3mm}
\centering
\scalebox{0.775}{
\begin{tabular}{c|ccccc} \hline
\multirow{2}{*}{Method} & \multicolumn{5}{c}{Pose detection error ratio (\%)} \\
& $0$ & $10$ & $20$ & $30$ & $40$ \\ \hline
ST-GCN~\cite{Sijie2018AAAI} & $83.3$ & $65.5$ & $63.5$ & $60.0$ & $52.0$ \\ 
Ours & $82.5$ & $80.8$ & $79.8$ & $78.5$ & $78.8$ \\ \hline
\end{tabular}
}
\label{tab:CompNoise}
\end{minipage}
\hspace{1.5mm}
\begin{minipage}[r]{0.25\textwidth}
\caption{Comparison of the domain shift on the RWF-2000 dataset.}
\vspace{-3mm}
\centering
\scalebox{0.775}{
\begin{tabular}{c|cc} \hline
& \multicolumn{2}{c}{Accuracy} \\
Method & Average & Variance \\ \hline
ST-GCN~\cite{Sijie2018AAAI} & $78.8$ & $0.81$ \\
Ours & $80.8$ & $0.08$ \\ \hline
\end{tabular}
\label{tab:ablation}
}
\end{minipage}
\hspace{1.5mm}
\begin{minipage}[r]{0.37\textwidth}
\centering
\includegraphics[clip, width=6.4cm]{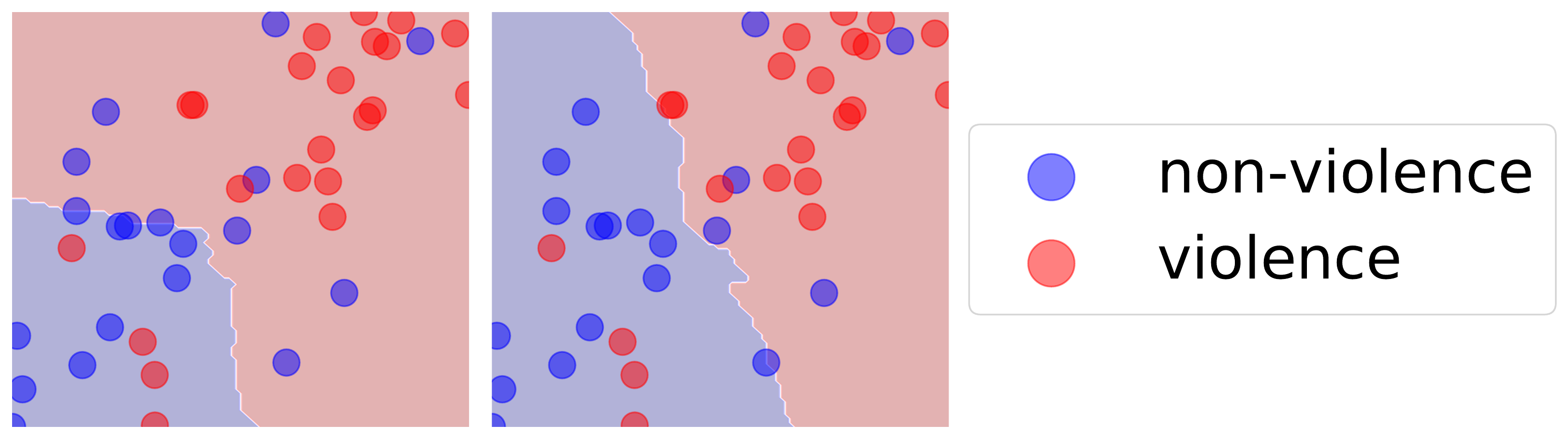}
\vspace{-7mm}
\captionof{figure}{Distribution of RWF-2000 samples in a 2D skeleton feature space compressed using t-SNE. The OoD score decision boundary (left) is shifted by the prompt-guided action score (right).}
\label{fig:t-sne2}
\end{minipage}
\end{table*}

\begin{table}[tb]
\caption{Comparison of the accuracy of the proposed method that uses different text prompt on the RWF-2000 dataset.}
\vspace{-3mm}
\centering
\scalebox{0.775}{
\begin{tabular}{c|cc} \hline
\multirow{2}{*}{Text prompt} & \multicolumn{2}{c}{Acc. (\%)} \\
& Only prompt & Full. \\ \hline
\multirow{2}{*}{\texttt{real fight}} & \multirow{2}{*}{$76.8$} & \multirow{2}{*}{$79.5$}\\
& & \\ \hline
\multirow{2}{*}{\texttt{punch or kick}} & \multirow{2}{*}{$76.5$} & \multirow{2}{*}{$79.3$}\\ 
& & \\ \hline
\multirow{2}{*}{\texttt{punch, kick or push}} & \multirow{2}{*}{$77.0$} & \multirow{2}{*}{$80.8$}\\ 
& & \\ \hline
\multirow{2}{*}{\shortstack{\\\texttt{punch, kick or push} \\ \texttt{related to violence}}} & \multirow{2}{*}{$79.0$} & \multirow{2}{*}{$81.5$}\\
& & \\ \hline
\multirow{2}{*}{\shortstack{\\\texttt{punch, kick or push} \\ \texttt{related to fighting}}} & \multirow{2}{*}{$80.0$} & \multirow{2}{*}{$82.5$}\\ 
& & \\ \hline
\end{tabular}
}
\label{tab:CompText}
\vspace{-1mm}
\end{table}

\cref{tab:RWF2000Acc,tab:K250Acc} summarize the anomaly action recognition accuracy of the proposed method as well as the state-of-the-art (SoTA) methods on the RWF-2000 and the Kinetics-250 datasets, respectively.
According to \cref{tab:RWF2000Acc}, the proposed prompt-guided framework (Ours) outperforms several previous supervised approaches in terms of accuracy, including PointNet++~\cite{Su2020ECCV}, DGCNN~\cite{Su2020ECCV}, and ST-GCN~\cite{Sijie2018AAAI}.
Although an inaccurate pose detector (PPN) is used in our method, its accuracy is also comparable to that of the SPIL~\cite{Su2020ECCV} by only 7 percentage points.
Additionally, \cref{tab:K250Acc} demonstrates that the accuracy of the proposed method (Ours) outperforms those of the SoTA unsupervised approaches.
These results of the proposed method are achieved without any DNN training in the target domain, although the previous methods take a time to train the DNNs.

Besides, the proposed fully-implemented anomaly score (Ours) outperforms its partial anomaly scores; the OoD score (Only OoD), and the prompt-guided action score (Only prompt), explained in \cref{sec:anomal,sec:user}, respectively.
According to \cref{tab:K250Acc}, the proposed method (Only OoD), which only uses the OoD score as the anomaly score, and the fully-implemented method (Ours), outperformed the previous unsupervised approaches.
As a result, the proposed method, which freezes the DNN weights during the training, can identify anomaly action in an unsupervised manner even if the text prompts are not provided.
Taking into account the aforementioned findings, the proposed method accomplishes the zero-shot anomaly action recognition that eliminates the target domain-dependent DNN training on normal samples, as described in \cref{sec:intro}.

Furthermore, the accuracy of the proposed method is enhanced by using the text prompt (Only prompt vs. Ours).
This result demonstrates that the proposed method reduces the misdetections that the normal action is identified as abnormal by supplementing the information of the abnormal or normal actions by using the text prompts (the second normal-sample limitation in \cref{sec:intro}).
\cref{fig:t-sne2} depicts the shifted decision boundary between abnormal and normal samples on the RWF-2000 dataset.
Besides, when comparing the accuracy of the proposed method (Only prompt), which uses only the prompt-guided action score, between \cref{tab:RWF2000Acc,tab:K250Acc}, the accuracy is severely degraded compared to the fully-implemented method (Ours) on the Kinetics-250, which is more notable than on the RWF-2000.
This is due to the proposed method only defining a few normal actions on the Kinetics-250 dataset without directly using the text prompts as abnormal actions.
As a result, the proposed method can detect abnormal behavior even when users define only normal actions.

\subsection{Ablation study} \label{sec:Ablation_Study}

\noindent \textbf{Comparison of robustness against skeleton detection and tracking errors.}

\cref{tab:CompNoise} compares the robustness against the skeleton detection errors (FPs, FNs, and tracking errors) described in \cref{sec:intro} between the proposed method and ST-GCN~\cite{Sijie2018AAAI} on the RWF-2000 dataset.
In this study, we synthesized three different types of skeleton detection errors: FPs, FNs, and tracking errors.
The FP errors were produced by adding the noise sampled from normal distribution to the two-dimensional joint coordinates.
By substituting the joint confidence score and joint coordinates with $0$ by a specific ratio, the FN errors were produced.
For instance, if the skeleton detection error ratio was $20\%$, we synthetically generated FP and FN errors to $20\%$ input joints and randomly switched their tracking indices by $60$ frames in $150$ frames of video for generating the tracking errors.
In comparison to the GNN-based supervised method~\cite{Sijie2018AAAI} in \cref{tab:CompNoise}, even if the skeleton error ratio rises, the accuracy of the proposed method does not degrade.

\noindent \textbf{Comparison of robustness against domain shift.}
We split the RWF-2000 training data into five subsets as different scenes and use each subset as a separate pattern that evaluates the methods.
\cref{tab:ablation} shows the average and variance of five accuracies for such five evaluations.
The variance of our method is clearly stable and represents robustness against the domain shift.

\noindent \textbf{Comparison of the accuracy against variations in text prompts.}
\cref{tab:CompText} presents the accuracy of the proposed method, which uses five different text prompts, with various anomaly scores.
Fully-implemented method (Full.) enhances the accuracy in a case that only the OoD score is employed as the anomaly score ($71.8\%$ in \cref{tab:RWF2000Acc}).
This demonstrates that using reasonable text prompts reduces misdetections of normal actions unobserved in the training phase.
Furthermore, for five text prompts, the accuracy of the prompt-guided action score is improved by using the OoD score (Only prompt vs. Full.).
As a result, the text prompt to identify abnormal actions, and the information gleaned from the normal data complement one another.

\section{Discussion}

\noindent \textbf{Generalization of feature extractor.}
The generalization of the proposed feature extractor naturally depends on the domain of its pretraining dataset.
We anticipate that this domain gap can be closed as a result of recent developments in dataset construction, which have made it possible to compile a sizable number of videos from Web and social media sources with captions.
Therefore, similar to the recent vision and language~\cite{Radford2021ICML,Changpinyo2021CVPR} and image anomaly detection~\cite{Rippel2021ICPR, Reiss2021CVPR} paradigms, more large-scale and generalizable representation learning can be conducted without manual annotation by employing a larger amount of captions and automatically extracted skeletons.

\noindent \textbf{Dependency on text prompt quality.}
The accuracy of the text prompt-guided zero-shot learning depends on the quality of text prompts and practically necessitates time-consuming prompt engineering.
Recent advances in prompt learning research have proposed context optimization~\cite{Zhou2022IJCV}, which has produced better results in the contexts of vision and language than zero-shot inference using hand-crafted prompts.
As a result, the anomaly action recognition accuracy benefits from not hand-crafted but learnable prompts and can automatically be improved.

\section{Conclusion}
This paper proposed a novel user prompt-guided zero-shot learning framework that can identify abnormal actions at the video level to address limitations in existing skeleton-based anomaly action recognition approaches.
Our core idea consists of three-fold:
(1) Leveraging a pretrained, target domain-invariant feature extractor that uses skeletons as inputs.
(2) Integrating a similarity score between skeleton features and user prompt embeddings aligned in the common space into the anomaly score.
(3) Creating a DNN architecture that is permutation-invariant and resistant to skeleton errors.
In the experiments, we tested the effectiveness of the proposed framework against the limitations.

{\small
\bibliographystyle{ieee_fullname}
\bibliography{paper}
}

\end{document}